# Probabilistic Impact Score Generation using Ktrain-BERT to Identify Hate Words from Twitter Discussions

Sourav Das, Prasanta Mandal, Sanjay Chatterji

*Department of Computer Science and Engineering,*
*Indian Institute of Information Technology, Kalyani*
*West Bengal, India*

**Abstract**
Social media has seen a worrying rise in hate speech in recent times. Branching to several distinct categories of cyberbullying, gender discrimination, or racism, the combined label for such derogatory content can be classified as toxic content in general. This paper presents experimentation with a Keras wrapped lightweight BERT model to successfully identify hate speech and predict probabilistic impact score for the same to extract the hateful words within sentences. The dataset used for this task is the Hate Speech and Offensive Content Detection (HASOC 2021) data from FIRE 2021 in English. Our system obtained a validation accuracy of 82.60%, with a maximum F1-Score of 82.68%. Subsequently, our predictive cases performed significantly well in generating impact scores for successful identification of the hate tweets as well as the hateful words from tweet pools.

## 1. Introduction

With the ever-evolving paradigms of global socio-economic issues, there has also been an increment in vast amounts of expressions and opinions on social networks in recent years. Depending on the context or topic, often the participating agents (humans) in the conversation can mutually agree or be polarized. Such contrast between the opinions and perceptions causes the tendency in people to get under the nerves of the other people involved in the discussion, even if it is by the help of hateful comments, or toxic content in general. The Cambridge Dictionary defines hate speech as "public speech that expresses hate or encourages violence towards a person based on something such as race, religion or sexual orientation". However, it could be unanimously agreed that political and financial viewpoints also play a significant role in modern hate content on social media.

In the need to counter the online hate content, the NLP research community is continuing prominent contributions and findings for the same [1, 2, 3]. However, hate speech is not always directly trackable, as it is dependent on topic and context [4]. Also, in a longer sequence of discussions, hate comments can occur as well we dissolve.

In this work, we focus on detecting hate speech and offensive content in English by exploratory data analysis with an ensemble transformer model.

The hate data HASOC produced by the FIRE 2021[1] [5], which are multilingual datasets for the Hate Speech and Offensive Content Identification shared task, but the datasets are available in separate fractions as per different languages (English, Hindi, and Marathi).

For the proposed research, we choose only the English dataset from HASOC Subtask 1. In order to shape the data for our proposed work, we introduce a set of extensive data analysis steps and reduce the data attributes only to the atomic columns. We split the train and test sets using linear Logistic Regression, and save the model. Next, we incorporate the ktrain BERT model [6], which is a lightweight BERT wrapped by the Tensorflow-Keras[2] library for low resource systems and a faster training phase. We extend the model further for several hyperparameters tuning, train and test visualization, and most importantly for probabilistic impact score generation from the trained model for any random sentences, to point out the hate speech factor (word) in those sentences. Finally, we produce the qualitative analysis by classification report from the saved model. We intend to make the code publicly available at Github[3].

The rest of the paper is organized as follows: Section 2 provides a detailed insight on the data and processing of the same. Section 3 describes the structural details of the selected model, with experimental and hyperparameter settings. Section 4 demonstrates the results with metrics and visualizations, probabilistic impact score generation for hate or positive tweets, and error analysis. Finally, Section 5 concludes the work with future directions.

## 2. Data Description

As mentioned earlier, our data for experimentation is the Hate Speech and Offensive Content Identification in English and Indo-Aryan Languages or HASOC 2021[4], provided by the Forum for Information Retrieval Evaluation (FIRE 2021) [7]. There are three separate sets of train and test corpora, respectively in English, Hindi, and Marathi. The datasets are primarily available for particular shared Subtask 1, which is further divided as subtask 1A for identifying hate and offensive contents from the tweets, and subtask 1B for discriminating between hate, profane, and offensive posts. We select the English corpus of Subtask 1 for our approach. The train set is a binary classification dataset, where the tweets contain the user id for each tweet, the tweets themselves in text format without the eradication of emojis, special characters, and external web links, and the classification category for each tweet. The columns are named "id", "text", and "task_1". The tweets are already annotated, and either labeled as HOF, i.e., Hate and Offensive Content, or NOT, i.e., Non-Hate-Offensive Content under the task_1 column. On the other hand, the test set contains tweets with respective user ids only. These tweets are not labeled and could either be used for hate content identification (in the form of HOF and NOT), or for system validation and hate tweet prediction. We showcase the detailed dataset information in Table 1.

| Data | Total Tweets | Labels | |
| --- | --- | --- | --- |
| | | HOF | NOT |
| HASOC 2021 Train (English) | 3844 | 2508 | 1976 |
| HASOC 2021 Test (English) | 1282 | NA | |

**Table 1**
HASOC 2021 Train and Test Dataset Details.

---

[1] http://fire.irsi.res.in/fire/2021/home
[2] https://www.tensorflow.org/api_docs/python/tf/keras
[3] https://github.com/SouravD-Me
[4] https://hasocfire.github.io/hasoc/2021/index.html

## 2.1 Preprocessing

To normalize the data in plain English texts, at first, we aim to eliminate the web links, abbreviations. We drop the additional columns not contributing to the determinant factor of the dataset. Next, we process the emojis and translate the labels (HOF or NOT) of the tweets into binary categories for better efficiency of the subsequent training phases. Here, hate speech can obviously be interpreted as negative content, while a generic tweet is a positive content. For that, we introduce two columns, neg and pos. the fundamental behind this is that we want to maintain an exact bipolarity of each tweet from the corpus. We maintain two labels parallelly, while joining the tweets with their respective HOF and NOT categories, and providing a score of 1 in a pos column if the tweet is non-hateful, or 1 in the neg column if the tweet contains hate speech. We show the head end of the preprocessed data shape in Figure 1.

|   | text | neg | pos |
|---|---|---|---|
| 0 | @wealth if you made it through this &&... | 1 | 0 |
| 1 | Technically that's still turning back the cloc... | 1 | 0 |
| 2 | @VMBJP @BJP4Bengal @BJP4India @narendramodi @J... | 0 | 1 |
| 3 | @krtoprak_yigit Soldier of Japan Who has    ... | 1 | 0 |
| 4 | @blueheartedly You'd be better off asking who ... | 1 | 0 |
| 5 | @ilyhiguchi | 1 | 0 |
| 6 | Why to blame only Modi or Government for this ... | 0 | 1 |
| 7 | The report on @TheLeadCNN that @clarissaward f... | 0 | 1 |
| 8 | @Chahal_Shekhar Sorry we won't, why can't your... | 1 | 0 |
| 9 | People are dying even in villages where there'... | 0 | 1 |

**Figure 1**: Head Columns of the Preprocessed Training Data.

## 3. Model

At first, we implement the already popularized Logistic Regression algorithm from scikit learn[5] to make a manual imbalanced partition of the train set itself. Since this is a classification task, LR has proved to be efficient in this genre of tasks [8]. LR maintains a uniform numerical regularization to normalize the stability of the training data, otherwise is also termed as the C value, i.e., the hyperparameter set. The downright advantage of deploying the LR model from scikit learn library is that it is fairly tuned, so the split (and training, if needed) needs little to no further optimization.

In order, we use the pre-trained BERT model wrapped in a Keras-Tensorflow library for efficiency in low code. This custom wrapper library is known as ktrain. BERT is already a widely renowned application. It is a bi-directional transformer that learns a language representation for pre-training. It uses the attention mechanism Transformer to learn the relationships between words in a text. Instead of having the model sequentially read the text input, the transformer encoder reads the entire text sequence at once. This characteristic makes it non-directional. But one major downside of such an acclaimed model is that classic BERT performs better than expected on many challenging tasks mainly due to its ability to handle large datasets, with Google's state-of-the-art computing resources, and the complexity of its pre-training tasks. This is not always beneficial for researchers with low resources or access. Hence, diluted and more comprehensive versions of BERT have emerged as the more popular choices for experimentations. A few of such models are DistilBERT [9], BERT-Base-Uncased [9], DeBERTa [10], etc.

---
[5] https://scikit-learn.org/stable/modules/generated/sklearn.linear_model.LogisticRegression.html#

A similar, but underutilized extension is the ktrain wrapper for BERT. It is not exclusively built for BERT, but the original repo is influenced by fastai[6] library, and also offers a wide range of other implementations for NBSVM [11], BiLSTM [12], and LDA [13]. It is also a potential library for image classification and graph neural networks. For dedicated NLP tasks, ktrain helps build and train neural networks for text classification, question answering, document summarization, named entity tagging, etc. It can also be used to estimate an optimal learning rate and schedule learning rates.

The first step involves loading and preprocessing data from different sources. Since ktrain is developed to work seamlessly with Keras, this step is quite similar to data loading in Keras models. It can be done in various forms, such as text, images, and graph data. At this stage, the model expects a preprocessor instance, where the preprocessor is loaded with the encapsulated feature set, pretraining length, and model range. In the second step, the model can be further customized by hyperparameter tuning, learning schedules interchanging, or by creating a custom one using tf.keras. Here we can employ the learning rate schedules such as one-cycle policy and SDGR. For our approach, we use the turnwise one-cycle policy, i.e., repeating a cycle each with 25 epochs for three consecutive times. After that, the model is automatically configured after inspecting the data. Next, we call the predictor method from the already learned model, and we provide instances (tweets) from the test data, or even random sentences to predict the category (HATE or NOT). Now, for the predicted set of sentences, the explain method can be called to generate the probabilistic scores to showcase the scores for each word in that sentence. A word with a higher score represents the higher negative impact and essentially turns out to be hate, offensive, or discriminatory content. Here the predictor log along with the model implementation could also be saved for later deployment. Our model with the proposed methodology for the experiment is represented in Figure 2.

### 3.1 Setup

We use Google Colab as our platform for experimentation. It introduces a native feature named a hardware accelerator for faster execution time, but with a limited resource threshold. Colab provides the options to choose from the dedicated graphical processor environment execution based on Tesla K80 GPUs or Google's Tensor Processing Units, developed for parallel neural computations simultaneously. Users can select any of the previously mentioned options, and for the record, any BERT-based model requires demanding resources. Since we worked with the ktrain wrapper, we select the notebook settings as GPU on a regular non-pro Colab account. We intend to utilize the available CUDA cores in a GPU environment for enhanced training and prediction time. We also connect to a hosted runtime with sufficient but abstract RAM and disk availability, as Google does not share the exact figurative information with the users.

### 3.2 Hyperparameter Settings

For the hate words classification from ktrain-BERT, at first, the pretrained BERT loaded with relatively fewer parameters to run on top of Keras. The ktrain-BERT has a pretraining length of 75, with a maximum feature set consideration, and set of words per batch for embedding. For word token sequencing, it has a ngram range of 3. The model loading also includes determining the column labels, so the model itself can pre-judge if the task is binary classification or multi-label classification. Then we call the get_learner function to place the already split data in the ktrain learner object. The batch size is kept as 16 for providing small training sets at a time. The learner object can also be invoked by creating a custom fit function. We initiate the fit function while keeping the learning rate $1e-5$ for 25 epochs. For the training process, we further tune the learning rate by implementing the ktrain find function, which provides an interactive customization scope into the learning schedule. We employ a Stochastic Gradient Restart schedule (SGDR) to ensure to maintain checkpoints in training and resume training if the notebook is paused, or the GPU access gets terminated. Now, the training metrics are fit to 8 epochs for 5 cycles, where only the best-performed batches are kept for entering into the next cycle, and the rest are discarded. Here we keep the learning rate as $1e-6$. Our total trainable parameters are 109,148,162.

---

[6] https://docs.fast.ai/

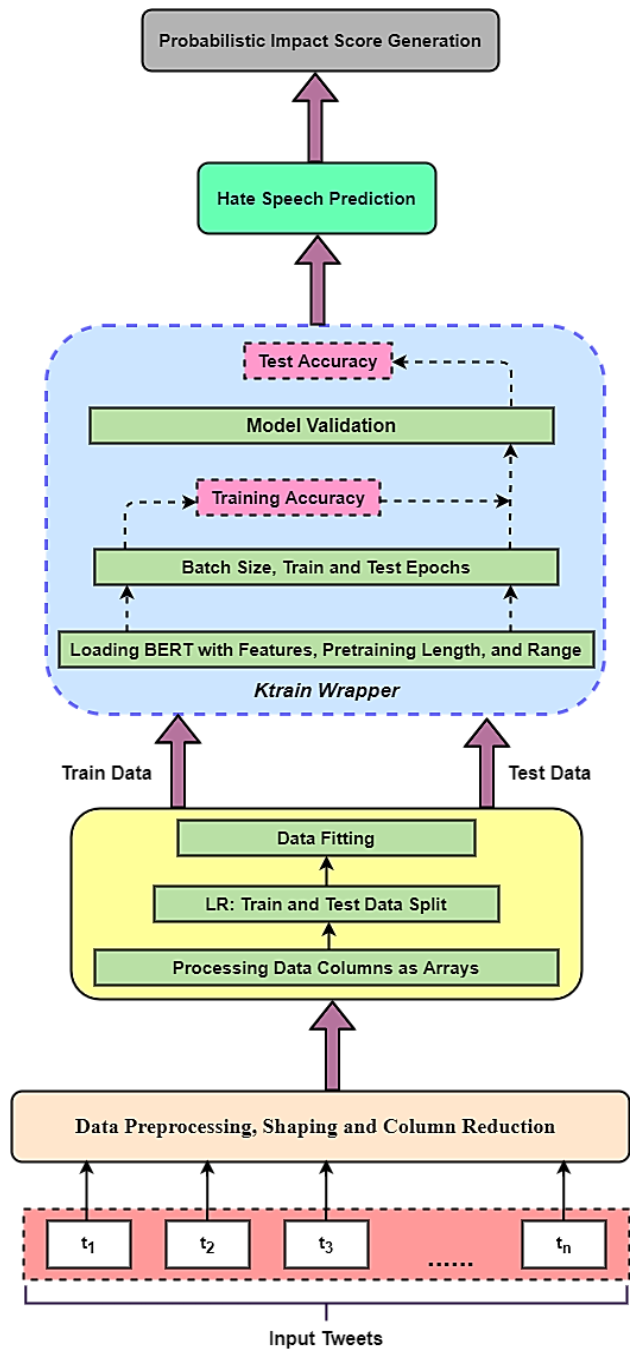

**Figure 2**: Methodology for Hate Speech Identification with Ktrain-BERT.

## 4. Results

The results of ktrain-BERT for dedicated hate words identification for English data are shown in Table 2. At next, we show the model shape for the first 5 layers out of a total of 12 layers with layer-wise structure in Table 3 to provide a better investigation of the model itself. We run the ktrain-BERT for 40 epochs in 5 consecutive cycles. The training vs. validation accuracy is plotted in Figure 3. It shows the incremental growth of the model performance w.r.t time during this phase. For English HASOC data, the Precision, F1-Score, and the highest training and validation accuracy obtained subsequently are 87.16%, 83.63%, 97.28%, and 82.60%. The classification report can be depicted in Figure 4.

| Language | Model | Identification Labels | Overall Metrics | | | |
|---|---|---|---|---|---|---|
| | | | Accuracy | Precision | Specificity | F1-Score |
| English | Ktrain-BERT | HOF (Hate Content) | 82.60% | 87.16% | 85.35% | 83.63% |
| | | NOT (Non-Hate Content) | | | | |

**Table 2**
Overall Performance Metrics for the Test Data.

| Layers | Layer Type | Parameters | Connected To |
|---|---|---|---|
| 1 | Input-Token and Segment (Input Layer) | NA | NA |
| | Embedding Token and Segment | 2344 | Input-Token-Segment & Embedding |
| 2 | Embedding Position, Dropout and Normalization | 57600, 1536, 2362368 | Embedding Layers |
| | Encoder 1 – Multihead Self Attention | 1536 | Encoder 1 Feedforward Add Layer |
| 3 | Encoder 3 – Feedforward MHA | 4722432 | Encoder 2 Feedforward Add Layer |
| 4 | Encoder 4 – Feedforward MHA | 2362368 | Encoder 3 Feedforward Add Layer |
| 5 | Encoder 5 – Feedforward MHA | 4722432 | Encoder 4 Feedforward Add Layer |

**Table 3**
Summary for Ktrain-BERT Model Used.

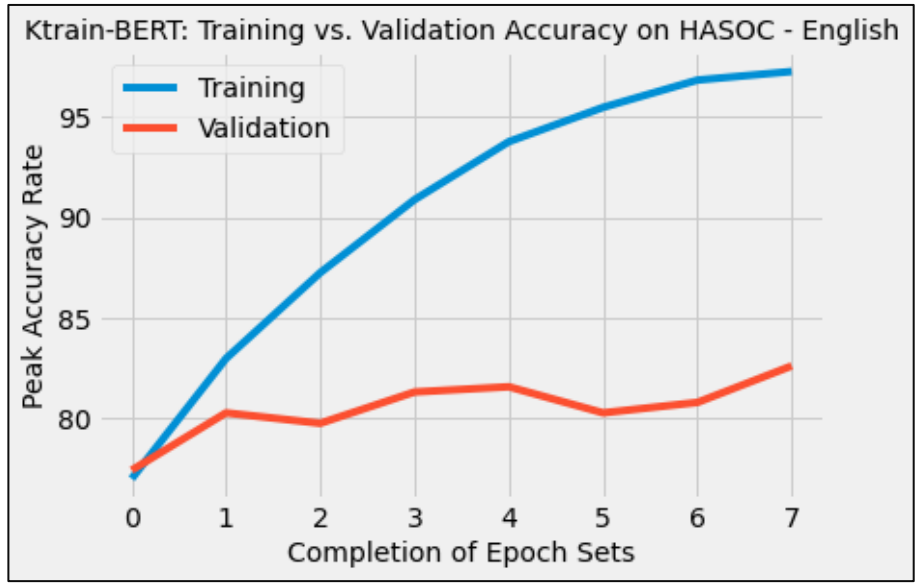

**Figure 3**: Training vs. Validation Accuracy by Ktrain-BERT Model.

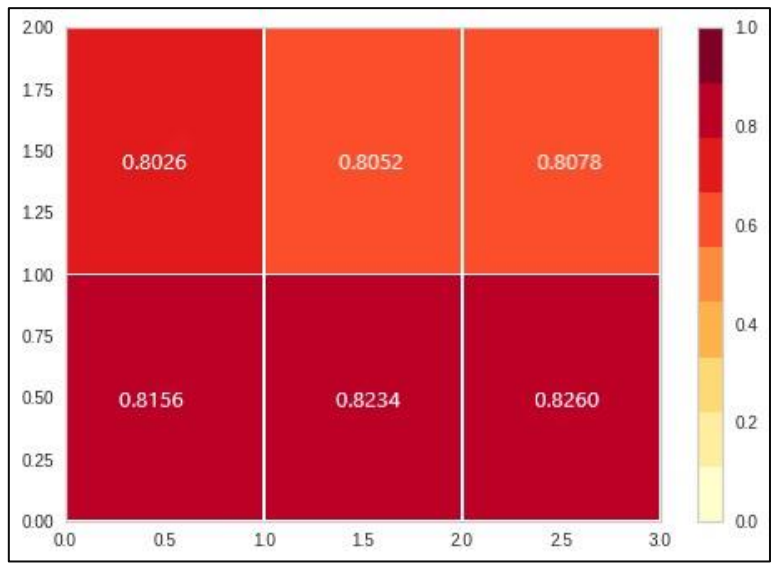

**Figure 4**: Classification Report for Validation Set.

## 4.1 Probabilistic Impact Score

After the validation phase, we choose random tweets both from the training and test data to explain the model on a predictive probabilistic scale. It is obtained by extending the trained set and validating the test data further by fusing the explain and predict function on the randomly fed tweets. We choose three tweets with hate content, and the other three with non-hateful content. The scores are normalized in a comparable scale, i.e., all the scores are in a non-negative scale, while quite naturally the higher scores represent more positive or negative impacts, reliant on the context of the concerning tweet itself. The results are shown here in Figures 5 to 10, with the impact score for each word within the sentences.

**y=pos** (probability **0.997**, score **5.796**) top features

| Contribution? | Feature |
|---|---|
| +6.574 | Highlighted in text (sum) |
| -0.778 | <BIAS> |

the razer blade 15 is really good this year!

**Figure 5**: Non-Hateful Tweet 1.

Tweet impact score breakdown: the - 0.767, razer - 0.073, blade - 1.143, 15 - 2.052, is - 2.336, **really** - **3.190**, **good** - **1.707**, this - 0.517, year - 1.590.

**y=pos** (probability **0.950**, score **2.941**) top features

| Contribution? | Feature |
|---|---|
| +3.403 | Highlighted in text (sum) |
| -0.461 | <BIAS> |

any daniel d. lewis movie is almost flawless.

**Figure 6**: Non-Hateful Tweet 2.

Tweet impact score breakdown: any - 0.539, daniel - 0.622, d - 0.922, lewis - 0.801, movie - 0.714, is - 0.223, **almost** - **1.595**, **flawless** - **2.604**.

**y=pos** (probability **0.844**, score **1.690**) top features

| Contribution? | Feature |
|---|---|
| +2.328 | Highlighted in text (sum) |
| -0.638 | <BIAS> |

real madrid has an excellent record in the champions league

**Figure 7**: Non-Hateful Tweet 3.

Tweet impact score breakdown: real - 0.259, madid - 1.032, has - 0.881, an - 0.002, **excellent** - **1.583**, record - 0.383, in - 0.171, the - 0.444, champions - 0.411, league - 0.652.

**y=neg** (probability **0.990**, score **-4.588**) top features

| Contribution? | Feature |
|---|---|
| +4.211 | Highlighted in text (sum) |
| +0.376 | <BIAS> |

ryzen is a winner, and you're a freak to think otherwise.

**Figure 8**: Hateful Tweet 1.

Tweet impact score breakdown: ryzen - 0.642, is - 0.977, a - 0.296, winner - 0.132, and - 0.036, you're - 0.808, a - 0.091, **freak - 1.920**, to - 0.353, think - 0.194, otherwise - 0.506.

y=neg (probability **0.997**, score **-5.660**) top features

| Contribution? | Feature |
|---|---|
| +5.230 | Highlighted in text (sum) |
| +0.429 | <BIAS> |

stop being a twat then

**Figure 9**: Hateful Tweet 2.

Tweet impact score breakdown: stop - 1.191, being - 1.630, a - 0.238, **twat - 2.383**, then - 0.818.

y=neg (probability **0.999**, score **-6.596**) top features

| Contribution? | Feature |
|---|---|
| +6.379 | Highlighted in text (sum) |
| +0.216 | <BIAS> |

you are hopeless. retire, wanker.

**Figure 10**: Hateful Tweet 3.

Tweet impact score breakdown: you - 2.082, are - 1.220, hopeless - 0.026, retire - 1.816, **wanker - 3.773**.

It is evident from the above demonstrations that the system has identified the impactful hate or non-hate words within the tweets in every instance. While it might be argued that the probabilistic impact score for the words "hopeless" or "retire" is relatively low to identify them as potential hate words, but the counter logic could be these words do not necessarily mean to use for derogatory comments always. These are heavily used for the generic purpose (e.g., "*It is sad that Jim is finally going to retire*", or "*I'm feeling hopeless since the beginning of the pandemic*"). Hence the impact on these words could not entirely determine the toxicity of a tweet or reply.

## 4.2 Error Analysis

It can be depicted from Figure 4 that even if the training classification accuracy constantly maintains a stable increment range over 90%, in fare to that the validation rather performed less significantly, with a bottleneck around 82%. The primary reason for this is the misclassification of several tweets during the validation phase. We showcase such one instance, where a tweet is like:

"*@ na ##ren ##dra ##mo ##di @ ami ##ts ##ha ##h @ ra ##hul ##gan ##dhi @ r ##g ##way ##ana ##do ##ffi ##ce its been long time holding my words…*"

Even after preprocessing and filtration, such tweets do not provide any particular insight for classification, hence these tweets fail to add any value to the overall validation and thereafter impact score generation process. The epoch performances are also relatively poor with higher loss where these tweets are retrieved and fed by the BERT for validating the classification. For the above-mentioned tweet, its ID is 139, and the epoch also suffers from a high loss value (loss: 7.77), in which it is originally entered within the batch. It has affected the overall validation accuracy also.

## 5. Conclusion and Future Work

In this work, we used a finetuned BERT wrapped in a custom Keras module for hate speech detection in English tweets. We obtained the highest training accuracy of 97.28% and a validation accuracy of 82.60%. Extending that work, we also identified the existence of hate or non-hate contents within concerned tweets using a probabilistic impact score generation for every word in a tweet, as well as with the whole tweet itself. For the further scope of this work, we want to develop a system capable of delivering the same for code-mixed tweets. Not only that, but we would also like to relate the hate speech detection to the tweet topics so that a separate scaling can be done for the topics; to determine which of the topics are more controversial than the others, and likely to stir up the hate speech, cyberbullying, and other forms of demeaning contents on the social networks.

**Declaration**

The hate speeches shown for the probabilistic impact score demonstrations are for experimental purposes only. These are real tweets and are collected from the training and/or test data. The authors DO NOT promote any form of hate content on social networks, and strongly condemn it.